\documentclass[12pt,conference,final,letterpaper,compsoc]{IEEEtran}

\usepackage{graphicx}
\usepackage{latexsym}

\newcommand{\bm}[1]{\mbox{\boldmath ${#1}$}}

\usepackage{amsmath}
\usepackage{algorithmic}

\usepackage{array}

\usepackage{fixltx2e}
\usepackage{dblfloatfix}

\usepackage{url}

\usepackage{fancyhdr}
\begin{document}
%
\title{Human Pose Estimation using Motion Priors and Ensemble Models}

\author{\IEEEauthorblockN{Norimichi Ukita}
\IEEEauthorblockA{Toyota Technological Institute\\
Email: ukita@toyota-ti.ac.jp}
}

\IEEEspecialpapernotice{(Invited Paper)}

\maketitle\thispagestyle{fancy}

\begin{abstract}
  Human pose estimation in images and videos is one of key
  technologies for realizing a variety of human activity recognition
  tasks (e.g., human-computer interaction, gesture recognition,
  surveillance, and video summarization).
  This paper presents two types of human pose estimation
  methodologies; 1) 3D human pose tracking using motion priors and 2)
  2D human pose estimation with ensemble modeling.
\end{abstract}



\section{Introduction}
\label{section:introduction}

Human pose estimation \cite{bib:poppe:survey,bib:sarafianos:survey}
has a large variety of applications such as action
recognition\cite{bib:thurau:action_recognition},
and biometrics\cite{bib:zhang:gait}.
Two different levels of human pose estimation problems have been
studied in Computer Vision. The first problem is 3D human pose
estimation, in which 3D joint positions/angles (e.g., $x$, $y$, and
$z$ coordinates of each body joint) are estimated. While the 3D human
pose is the complete representation of a 3D human body, the 2D
configuration of body joints observed in an image is also useful for
understanding human activities. Therefore, 2D human pose estimation,
the second problem, is also one of the hot research topics.

3D pose estimation is more difficult than 2D pose estimation.
However, several additional cues, such as multi-viewpoint images and
temporal smoothness in videos, allow us to resolve this problem.
Among these cues, we focus on a motion prior captured in a video.
To estimate complex dynamic human poses, the motion prior is widely used
for accuracy and robustness. Most recent works obtain it
from sample sequences of real human motions.
Several kinds of actions, such as walking and running, are recorded in
motion datasets that are widely used 
in Computer Vision\cite{bib:black:humaneva} and
Graphics\cite{bib:cmu:mocap,bib:mit:mocap} communities.
The motion model of each action can be used for pose tracking in that
action.

Unlike 3D human pose tracking in videos, motion
priors cannot be employed for 2D human pose estimation in still
images.
Instead of motion priors, other additional schemes are used for 2D
human pose estimation.
We focus on a limited variety of possible body poses depending on the
scenario (e.g., human activity).
In basic methods for pose estimation, the appearance features and body
joint distributions of a human body are modeled in a training process.
Human pose estimation is challenging due to the wide variety of
appearances and joint distributions.
One way to alleviate the complexity is to cluster a training dataset
so that a set of ensemble models can be learned.
Reducing the variation within each subset facilitates learning the
ensemble model to accurately estimate the joint locations under a
particular pose configuration.



\section{3D Human Pose Tracking using Motion Priors}
\label{section:3d}

\begin{figure}[t]
  \centering
  \begin{minipage}{0.35\columnwidth}
    \centering
    \includegraphics[width=0.75\columnwidth]{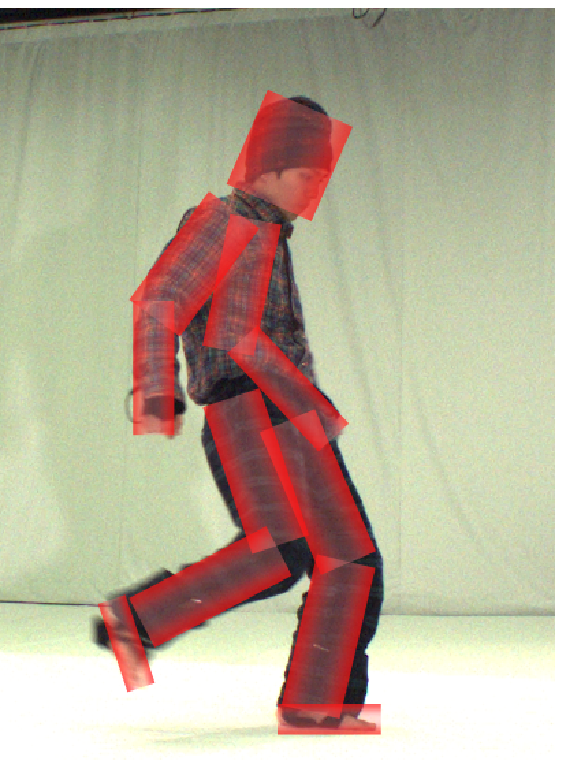}\\
    {\small (a) Model matching: generative approach}
  \end{minipage}~
  \begin{minipage}{0.63\columnwidth}
    \centering
    \includegraphics[width=\columnwidth]{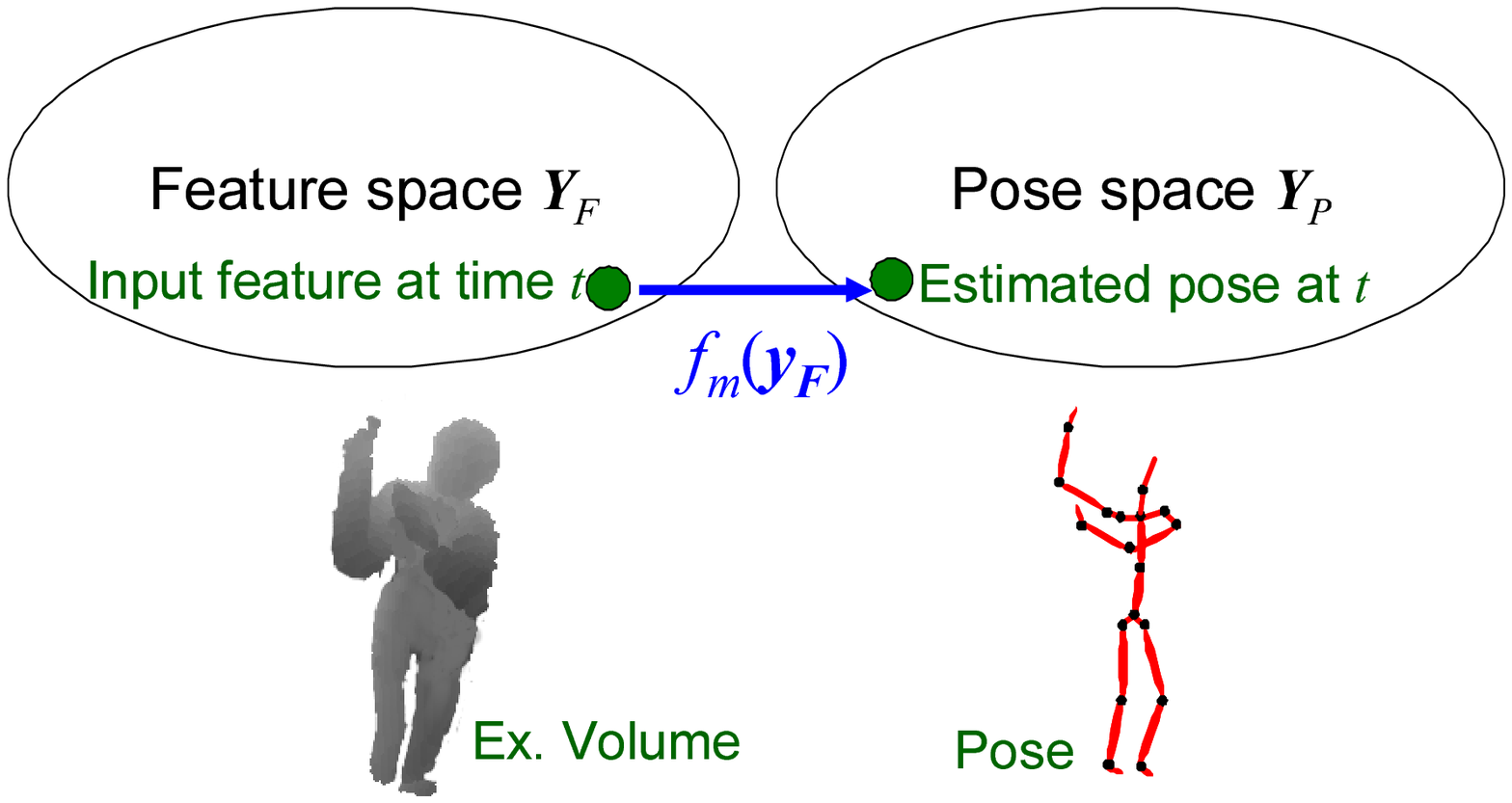}
    (b) Feature-to-pose mapping: discriminative approach.
  \end{minipage}
  \caption{Pose parameter estimation from an observed image}
  \label{fig:pose_regression}
\end{figure}

In 1990s, most human pose estimation methods were based on generative
approaches where pose parameters (i.e., joint positions) are optimized
so that each body part model overlaps its image as shown in Figure
\ref{fig:pose_regression} (a).
Recent advances in machine learning allow us to apply discriminative
approaches to human pose estimation.
While most of our methods are based on a feature-to-pose regression
\cite{bib:ukita:vh_constraints,DBLP:conf/iccv/GirshickSKCF11,VNect_SIGGRAPH2017},
body-part segmentation based on classification
\cite{DBLP:conf/cvpr/ShottonFCSFMKB11} can be also useful.
In an example shown in Figure \ref{fig:pose_regression} (b), a feature
vector $\bm{y}_{F}$ is computed from the 3D volume of a
target person, and a regression function
$\bm{y}_{P} = f_{m}(\bm{y}_{F})$ is trained to obtain a set
of pose parameters $\bm{y}_{P}$.

In generative approaches, most methods assume that the 3D shape of a
person is represented as an articulated object consisting of rigid
parts. This is because it is difficult to efficiently represent the
flexible shape of loose-fitting clothing by 3D model deformation
\cite{bib:black:naked}.
Feature-to-pose regression approaches, on the other hand, can easily
represent such large shape deformation
\cite{bib:ukita:vh_constraints}.

For successful discriminative approaches, we have to acquire mapping
functions between image/shape features and body pose parameters. In
order to cope with high dimensionality of the image/shape features,
various low dimensional features that are robust to observation noise
have been proposed
\cite{bib:bustos:3d_model}.
General dimensionality reduction algorithms are also applicable to
this feature extraction; linear algorithms such as PCA, nonlinear
algorithms such as Locally Linear Embedding and
Isomap, and probabilistic nonlinear
embedding methods such as Gaussian Process Latent variable Models,
GPLVM \cite{bib:lawrence:gplvm}.

\begin{figure}[t]
  \centering
  \includegraphics[width=\columnwidth]{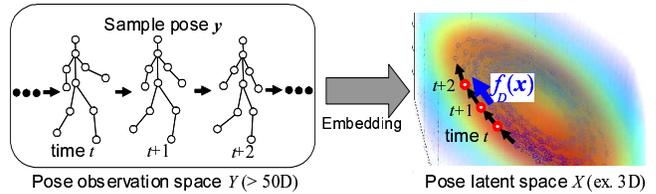}
  \caption{Pose space $Y$, its latent space $X$,
    and mapping functions between and within them.
    Circles and arrows in $X$ depict latent variables and temporal
    mapping, $f_{D}(\bm{x})$, respectively.}
  \label{fig:embedding}
\end{figure}

For tracking a complex pose sequence, motion priors are useful.
Dynamic and complex human motions can be represented by parametric
models and/or 
instance-based models.
In most motion models, the dimension of a feature space is reduced for
improving model generalization. In a toy example shown in Figure
\ref{fig:embedding}, motion dynamics is modeled in a low-dimensional
latent space $X$ obtained from it original high-dimensional features
(i.e., sample pose sequences $\bm{y}$).

\subsection{Human Pose under Clothing}
\label{subsection:complex}

To estimate complex dynamic poses, pose tracking using motion priors
and multiview images is more effective than pose detection from a
unidirectional view.

\begin{figure}[t]
  \centering
  \includegraphics[width=\columnwidth]{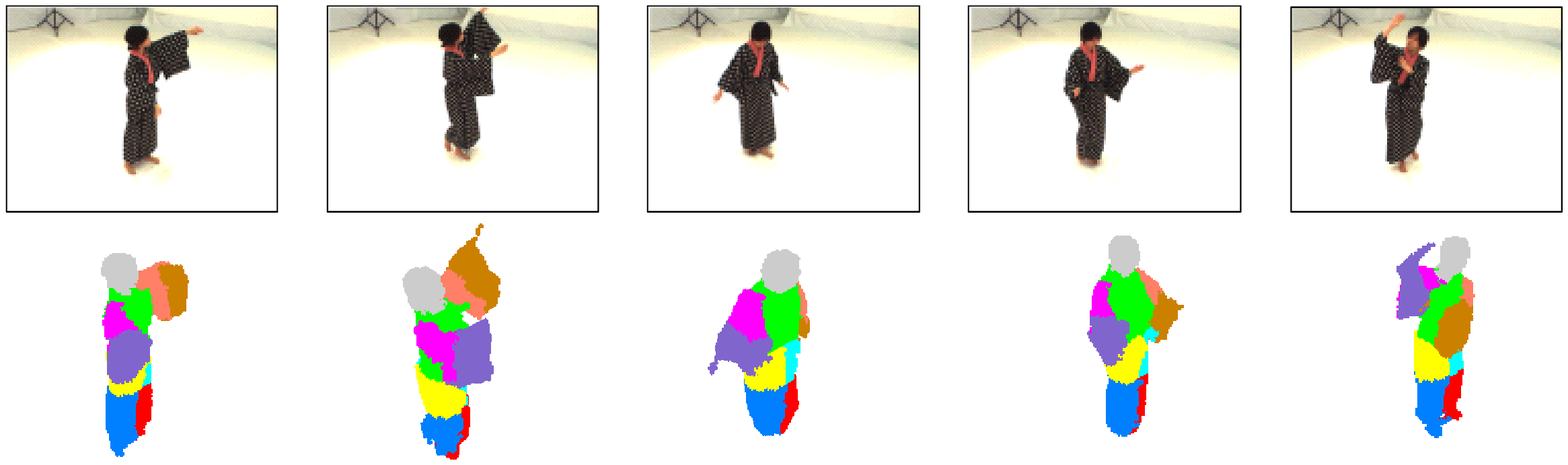}
  \caption{Volume refinement and body-part labeling. 1st row:
    observed images, 2nd row: Body-part labeled voxels.}
  \label{fig:eccv08results}

  \vspace*{6mm}

  \centering
  \includegraphics[width=\columnwidth]{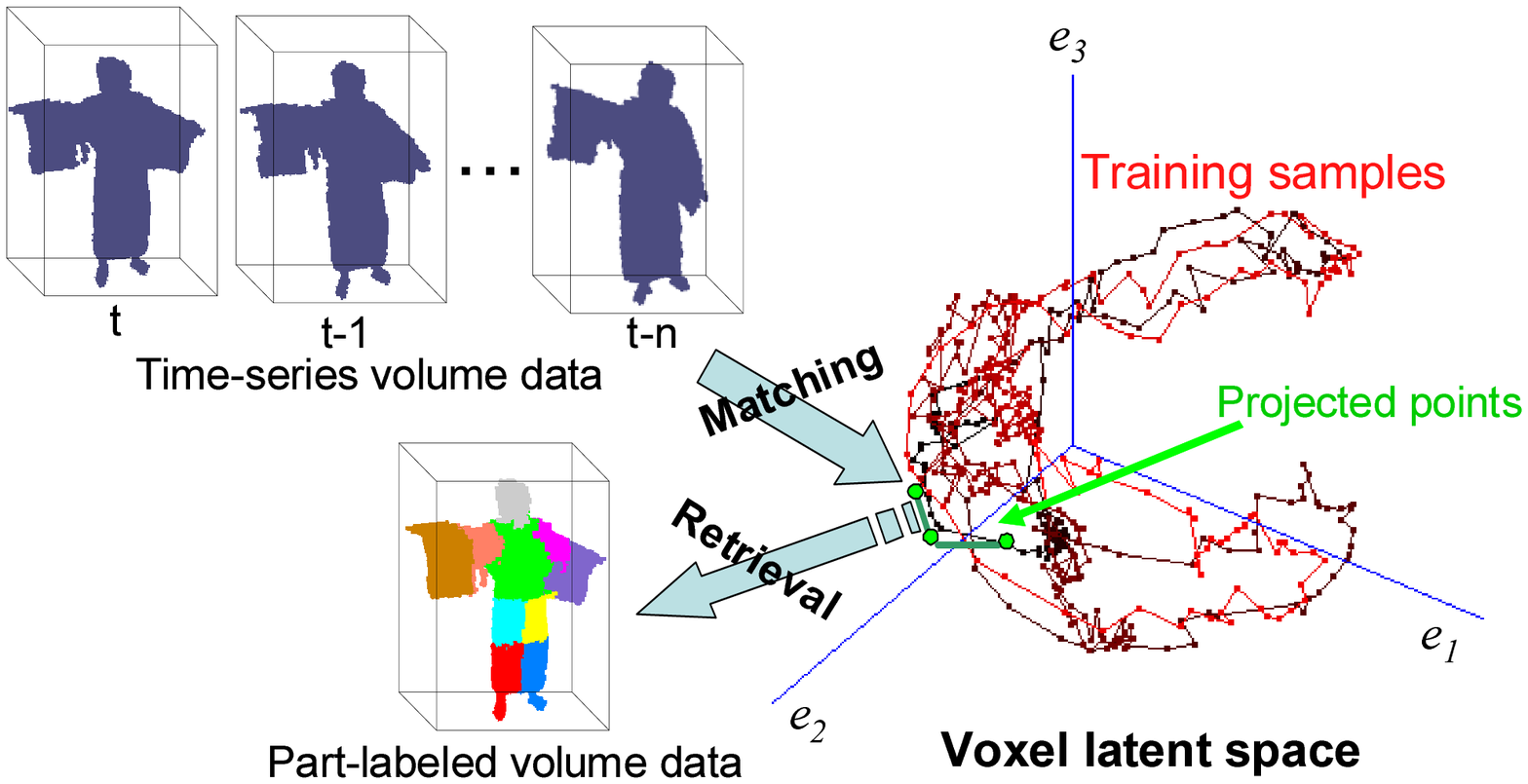}
  \caption{Matching time-series volumes in their latent space and
    retrieving body-part labeled voxels from the matching result.}
  \label{fig:eccv08}
\end{figure}

In \cite{bib:ukita:part}, our goal is 3D body-part segmentation in a
reconstructed volume of a human body wearing loose-fitting clothing,
as shown in Figure \ref{fig:eccv08results}.
While the similar goal is achieved by framewise body-part labeling
using a huge number of training data generated by computer graphics in
\cite{DBLP:conf/cvpr/ShottonFCSFMKB11}, our method uses temporal
matching with real-image training data.
A set of time-series target volumes, which is acquired by a slow but
sophisticated 3D reconstruction algorithm, with body-part labels is
learned in advance. The time-series sample volumes are learned using
PCA and stored as the manifolds in the eigenspace, as shown in ``Voxel
latent space'' in Figure \ref{fig:eccv08}. Each input volume
reconstructed online is projected into the eigenspace and compared
with the manifolds in order to find similar high-precision samples
with body-part labels.
%
%


\begin{figure}[t]
  \centering
  \includegraphics[width=\columnwidth]{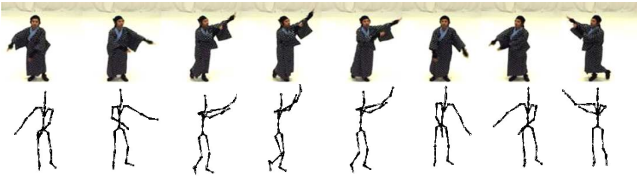}
  \caption{Human volume tracking with error handling and pose
    estimation. 1st row: observed images, 2nd row: Estimated poses.}
  \label{fig:iccv09results}
\end{figure}

%
In \cite{bib:ukita:vh_constraints}, complex motion priors of a target
shape are modeled explicitly with the temporal mapping function
defined in a low-dimensional space
modeled by probabilistic non-linear embedding, i.e., Gaussian Process
Dynamical Models, GPDM \cite{bib:fleet:gpdm}.
We also obtain the low-dimensional space of a human body pose
and train a mapping function from the volume latent space to the pose
latent space.
Since all of these mapping function are defined by Gaussian process
regressions, these functions are generalized in contrast to shape
matching/retrieval used in \cite{bib:ukita:part}.
For further robustness, reconstructed volume tracking is achieved by
particle filtering, where the likelihood between the reconstructed
volume
and each particle
in the volume latent space is computed in the volume latent space.
The likelihood-weighted mean
of all particles is regarded as the target volume at this time step.
For the next time step, all particles are shifted using the temporal
mapping function defined by GPDM.
Finally, the estimated refined volume is mapped to the pose latent
space in order to compute the body pose of the target person.
As shown in Figure \ref{fig:iccv09results}, we can estimate the body
poses under loose-fitting clothing.

\subsection{Pose Tracking in Multiple Actions}
\label{subsection:actions}

Motion priors can be modeled and applicable to pose tracking as
described in Section \ref{subsection:complex}.
Training data of motion priors can be obtained from various motion
datasets \cite{bib:black:humaneva,bib:cmu:mocap,bib:mit:mocap}.
Different kinds of actions (e.g.  walking, jogging, dance) are
recorded independently in these datasets. The motion model of {\em
  each action} can be leveraged for analyzing that action.
Different actions are smoothly transited from one to another (e.g.
from walking to jogging) in a natural scenario, while they are
recorded independently in motion datasets.
For efficiently using motion priors of {\em multiple actions} in such
natural scenarios,
we proposed two kinds of motion modeling schemes where different
training sequences are connected via transitions paths.
While motion transitions among training sequences have been proposed
for graphics animations
\cite{bib:gleicher:motion_graphs,bib:safonova:motion_graphs}, we
optimize the transition paths for vision-based tracking problems.

\begin{figure}[t]
  \centering
   \includegraphics[width=0.48\columnwidth]{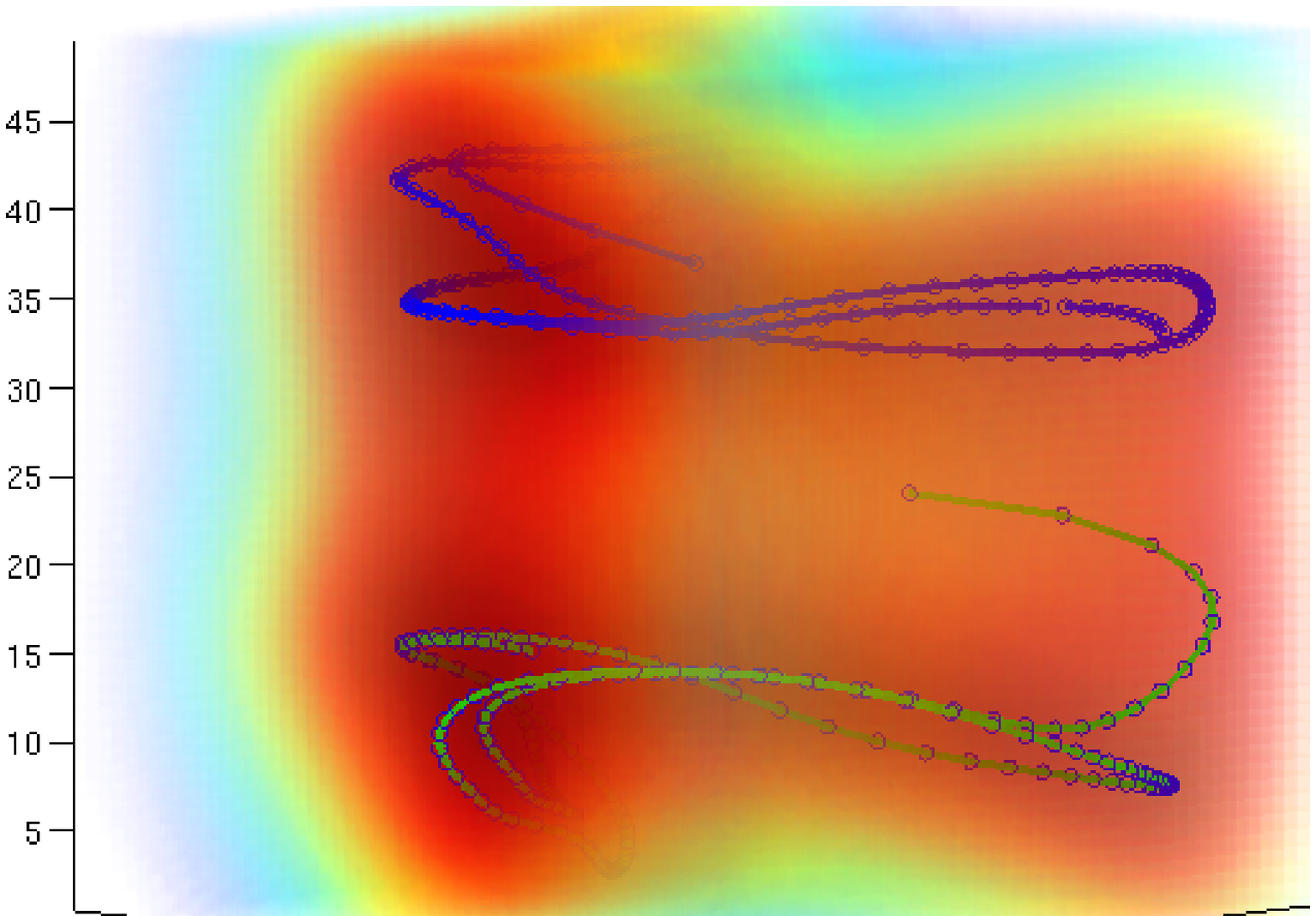}~
   \includegraphics[width=0.48\columnwidth]{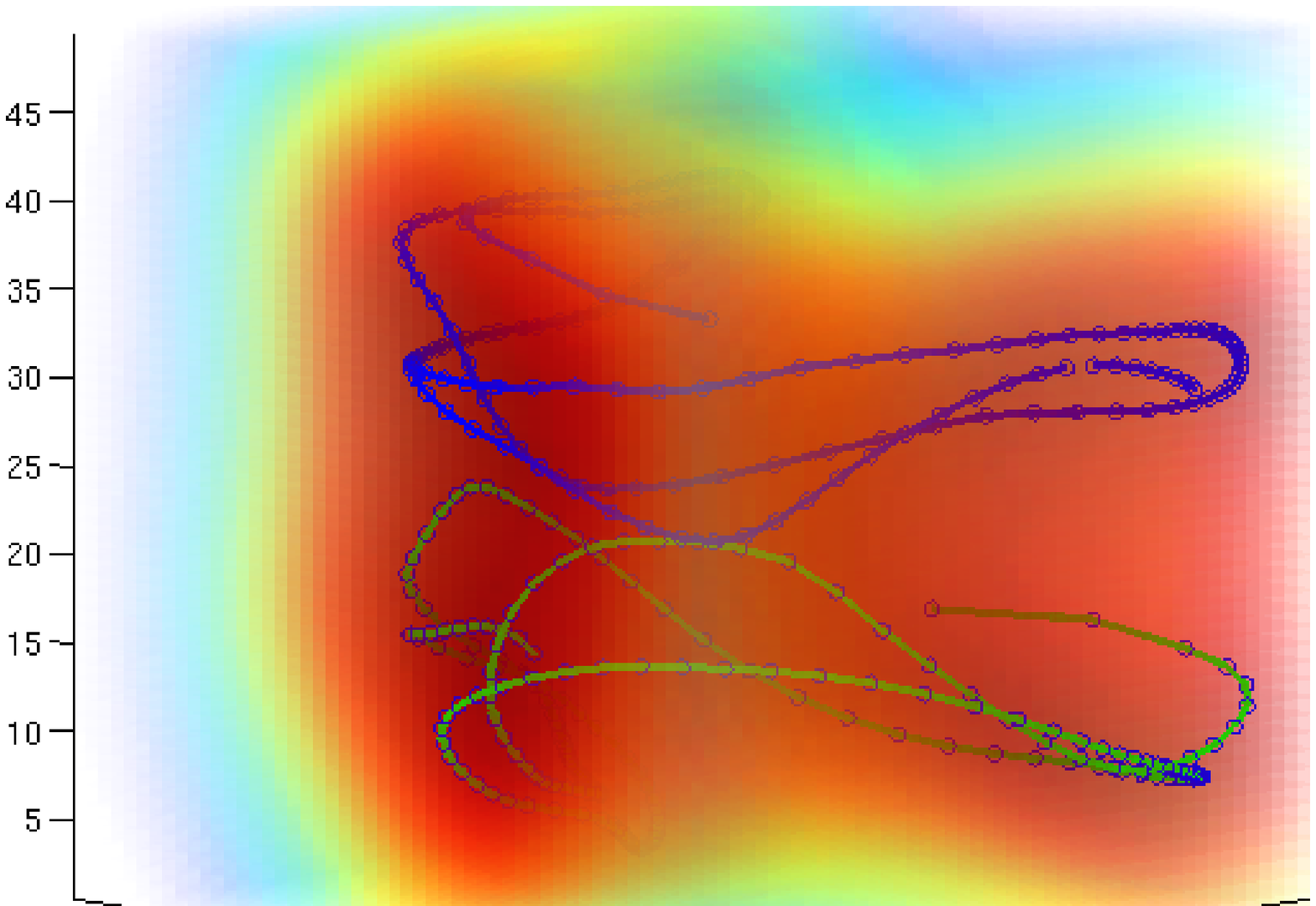}\\
   (a) Standard GPDM \hspace*{12mm} (b) TC GPDM
   \caption{Latent models obtained by GPDMs; Left: GPDM, Right:
     Topologically-constrained GPDM. The blue and green arrows show
     dance1 and dance2 sequences, respectively.}
   \label{fig:unified}
\end{figure}

\begin{figure}[t]
  \centering
  \includegraphics[width=\columnwidth]{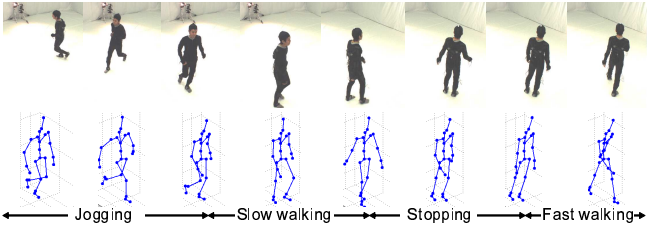}
  \caption{Pose tracking with multiple actions. 1st row: observed
    images, 2nd row: Estimated poses.}
  \label{fig:eccv10results}
\end{figure}

{\bf Unified model:} The motions of all actions are modeled in a
unified latent model \cite{DBLP:journals/cviu/UkitaK12}.
In order to optimize the latent model with kinematically-realistic
transitions between different actions, topologically-constrained
modeling \cite{bib:lawrence:topological} is used with constraints such
that each transition path connects two different action sequences as
smooth and short as possible, as shown in Figure \ref{fig:unified}.
Then these two different sequences are connected via synthesized paths
for smooth action transitions.
With this model, error in 3D body-joint localization during action
transitions decreases 49 \% on average in contrast to a latent model
produced by GPDM with no transition paths.

{\bf Separate models:} The motion of each action is modeled in its
independent latent model \cite{DBLP:journals/ivc/Ukita13}. Such
independent modeling of action-specific motions allows us 1) to
optimize each model in accordance with only its respective motion and
2) to improve the scalability of the models.
For robust tracking with the multiple models, particle filtering is
employed so that particles are distributed simultaneously in the
models. Efficient use of the particles can be achieved by locating
many particles in the model corresponding to an action that is
currently observed.
For transferring the particles among the models in quick response to
changes in the action, transition paths are synthesized between the
different models.
The effectiveness of the proposed models is validated with several
datasets. Figure \ref{fig:eccv10results} shows tracking results in an
image sequence including six gait actions; only four of them are shown
in this example.
Compared with independent models with no transition paths, error in
body-joint localization decreases 20 \% on average in our proposed
models.


\section{2D Human Pose Estimation with Ensemble Modeling}
\label{section:2d}

Pictorial structure models (PSMs)
\cite{DBLP:journals/ijcv/FelzenszwalbH05} have been extensively
applied to 2D human pose estimation because of their ability for
efficient and global optimization.
PSMs can be augmented by discriminative training
\cite{DBLP:journals/pami/FelzenszwalbGMR10,DBLP:journals/pami/YangR13}.
Several extensions have been proposed to improve PSMs, including
coarse-to-fine modeling
\cite{DBLP:conf/eccv/SappTT10}, appearance
modeling using region segmentation \cite{bib:ukita:part_feature},
occlusion-robust modeling \cite{bib:ukita:occlusion}, and appearance
learning between parts \cite{DBLP:conf/cvpr/Ukita12,Chen_NIPS14}.
While global optimality of the PSM is attractive, its ability to
represent complex relations among joint locations is limited compared
to deep neural networks.

Convolutional neural networks (ConvNets) have recently been applied to
pose estimation.
A ConvNet can directly estimate the joint
locations~\cite{toshev2014deeppose} or estimate the pixel-wise
likelihood of each joint location as a
heatmap~\cite{tompson2014joint}.
Recent approaches explore sequential structured estimation to
iteratively estimate the part
locations~\cite{carreira2016human,wei2016cpm}.


\subsection{Action-specific Pose Models}
\label{subsection:asm}

\begin{figure}[t]
  \centering
  \includegraphics[width=\columnwidth]{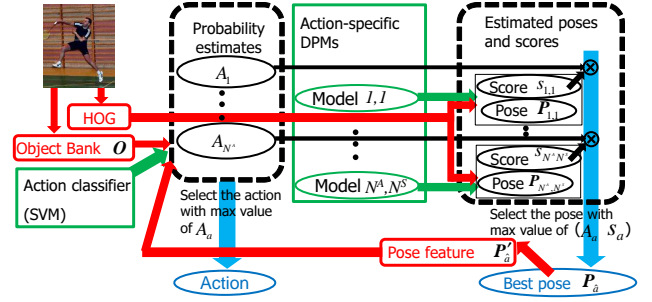}
  \caption{Action-specific pose models and its combination with action
  recognition.}
  \label{fig:action_dpm}
\end{figure}

\begin{figure}[t]
  \centering
  \includegraphics[width=\columnwidth]{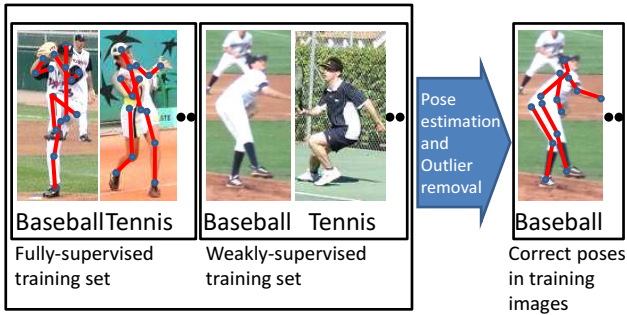}
  \caption{Weakly-supervised learning with action-specific 
    models.}
  \label{fig:weak_training}
\end{figure}

We can roughly understand a human activity from a human pose.
That is, human pose and action are mutually related to each other.
Based on this idea, action classification has been achieved by pose
matching (e.g.\ 2D pose-based matching
\cite{DBLP:journals/pami/WangM11} in videos and view-invariant 3D pose
matching in videos \cite{DBLP:conf/iccv/SinghN11}).
In an opposite manner, for pose tracking in videos, action-specific
model selection has been also achieved (e.g., efficient particle
distribution in pose models \cite{DBLP:journals/ivc/Ukita13} and
unified multi-action modeling \cite{DBLP:journals/cviu/UkitaK12}).

With the aforementioned mutual augmentation,
in \cite{bib:ukita:iterative}, we proposed an iterative scheme between
action classification and pose estimation in still images,
as shown in Figure
\ref{fig:action_dpm}.
Initial action classification is achieved only by global image
features that consist of the responses of various object filters.  The
classification likelihood of each action (``Probability estimates'' in
the figure) weights human poses estimated by the pose models of
multiple action classes (``Model'' in the figure).  Such
action-specific pose models allow us to robustly identify a human pose
under the assumption that similar poses are observed in each action.
From the estimated pose (``Best pose'' in the figure), pose features
are extracted and used with global image features for
re-classification.
With this scheme, pose accuracy increases 11.3\% compared with
the base model \cite{Chen_NIPS14}.

The aforementioned action-specific pose models can be employed for
weakly-supervised learning.
Assume that most training images have only the action label of a
person of interest (i.e., ``Weakly-supervised training set'' in Figure
\ref{fig:weak_training}), while some training images have also a human
pose annotation (i.e., ``Fully-supervised training set'' in the
figure).
By utilizing the fact that the pose features of the same action make
clusters, we estimate a human pose in each weakly-supervised image and
classify whether or not the estimated pose is correct.
The correctly-estimated pose is employed with its image for
re-training the pose model of its corresponding action.
The detail of this work will be introduced in the talk.

\subsection{Pose Modeling with Pose Similarity}
\label{subsection:pme}

\begin{figure}[t]
  \centering
  \includegraphics[width=\columnwidth]{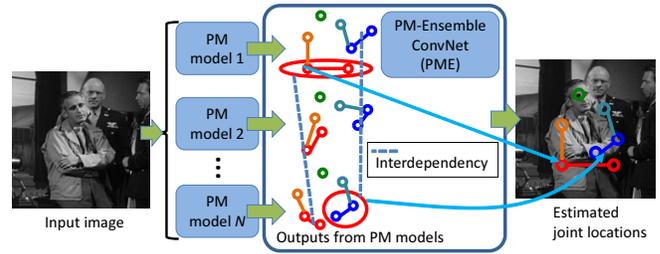}
  \caption{Pose outputs of heterogeneous pose models (PM models) are
    merged in accordance with interdependency among the pose outputs
    by a deep neural network.}
  \label{fig:pem}
\end{figure}

Unlike action-specific models, pose models can be clustered also
depending on pose similarity independently of human actions.
Figure \ref{fig:pem} shows the basic scheme in pose inference in this
approach. Before this inference, heterogeneous pose estimation models
(``PM model'' in the figure) are trained from the full set of training
images.
In the inference, all pose estimation models estimate their own
outputs independently. Then all the pose outputs are merged in order
to obtain the final output (``Estimated joint locations'' in the
figure).
This pose mergence is achieved by a huge deep neural network for
capturing complex interdependency among noisy and ambiguous output
poses.
This is a major differnece from pose selection from multiple
candidates \cite{imc}.
The detail of this work will be also introduced in the talk.


\section{Conclusion}
\label{section:conclusion}

All pose estimation models presented in this paper
utilize motion priors or ensemble modeling for improving the
performance.
For future work, complex and huge data representation using neural
networks and semi/weakly-supervised approaches should be important in
order to represent a huge variety of human activities.



\end{document}